\pdfoutput=1

\documentclass[11pt]{article}

\usepackage[final]{acl}

\usepackage{times}
\usepackage{latexsym}

\usepackage[T1]{fontenc}

\usepackage[utf8]{inputenc}

\usepackage{microtype}

\usepackage{inconsolata}

\usepackage{graphicx}

\usepackage{microtype}

\usepackage{booktabs}  
\usepackage{graphicx}
\usepackage{arydshln}

\usepackage{multirow}
\usepackage{float}
\usepackage{amsmath}
\usepackage{mismath}
\usepackage{hyperref}
\usepackage[many]{tcolorbox}
\usepackage{amssymb}
\usepackage{mathtools}
\usepackage{amsthm}
\usepackage{pifont}
\usepackage{enumitem}

\newcommand{\cmark}{\ding{51}}
\newcommand{\xmark}{\ding{55}}

\makeatletter
\def\eg{\textit{e.g.}} 
\def\ie{\textit{i.e.}} 
 
\def\etc{\textit{etc.}}

\makeatother

\newcommand{\keypoint}[1]{\noindent \textbf{#1}}

\newcommand{\ours}{Learnable Calibration}

\usepackage{subcaption}

\usepackage{algorithm}
\usepackage[noend]{algpseudocode}

\usepackage[english]{babel} 

\usepackage{tikz}
\usetikzlibrary{decorations.pathreplacing,calc}

\tikzset{brace/.style={decorate, decoration={brace}},
 brace mirrored/.style={decorate, decoration={brace,mirror}},
}

\newcounter{brace}
\newcounter{arrow}
\title{Efficient Compositional Multi-tasking for\\ On-device Large Language Models}

\author{
 \textbf{Ondrej Bohdal\textsuperscript{1}},
 \textbf{Mete Ozay\textsuperscript{1}},
 \textbf{Jijoong Moon\textsuperscript{2}},\\
 \textbf{Kyeng-Hun Lee\textsuperscript{2}},
 \textbf{Hyeonmok Ko\textsuperscript{2}},
 \textbf{Umberto Michieli\textsuperscript{1}}
\\
\\
 \textsuperscript{1}Samsung R\&D Institute UK, United Kingdom,
 \textsuperscript{2}Samsung Research, South Korea
\\
 \small{
   \textbf{Correspondence:} \href{mailto:o.bohdal.1@samsung.com}{o.bohdal.1@samsung.com}
 }
}

\begin{document}
\maketitle
\begin{abstract}
Adapter parameters provide a mechanism to modify the behavior of machine learning models and have gained significant popularity in the context of large language models (LLMs) and generative AI. These parameters can be merged to support multiple tasks via a process known as task merging. However, prior work on merging in LLMs, particularly in natural language processing, has been limited to scenarios where each test example addresses only a single task. In this paper, we focus on on-device settings and study the problem of text-based compositional multi-tasking, where each test example involves the simultaneous execution of multiple tasks. For instance, generating a translated summary of a long text requires solving both translation and summarization tasks concurrently. To facilitate research in this setting, we propose a benchmark comprising four practically relevant compositional tasks. We also present an efficient method (Learnable Calibration) tailored for on-device applications, where computational resources are limited, emphasizing the need for solutions that are both resource-efficient and high-performing. Our contributions lay the groundwork for advancing the capabilities of LLMs in real-world multi-tasking scenarios, expanding their applicability to complex, resource-constrained use cases.
Project page: \url{https://ondrejbohdal.github.io/CompositionalMultitaskingLLMs}.
\end{abstract}

\section{Introduction}
\label{sec:introduction}

\begin{figure}[ht]
\vskip 0.2in
\begin{center}
\centerline{\includegraphics[trim=1cm 0cm 0cm 0cm, clip, width=\linewidth]{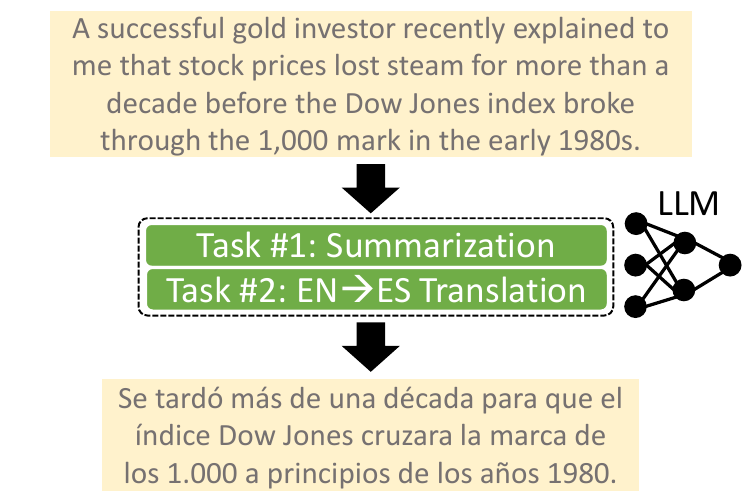}}
\caption{\textbf{Compositional multi-tasking} involves performing multiple tasks simultaneously, such as summarization and translation. The challenge lies in executing all tasks jointly within a single inference pass for optimal efficiency, rather than performing them separately through multiple inferences.
}
\label{fig:compositional}
\end{center}
\vskip -0.3in
\end{figure}

Large language models (LLMs) have revolutionized natural language processing (NLP) by demonstrating remarkable capabilities across a wide range of text-based tasks \cite{zhao2023survey,minaee2024large}. 
As foundational models, LLMs can be fine-tuned to excel in various downstream applications, such as question answering \cite{sticha2024qa}, summarization \cite{liu2023learning}, translation \cite{zhu2023multilingual}, and beyond \cite{shu2024rewritelm,severyn2021grammar}.
A common approach to fine-tuning or adapting LLMs is through parameter-efficient fine-tuning (PEFT), where relatively few additional parameters, often referred to as adapters, are inserted into the model \cite{han2024parameter,ding2022delta}.
Among these methods, Low-Rank Adapters (LoRA) \cite{hu2021lora} have gained prominence due to their efficiency and ease of integration. 
LoRA trains low-rank matrices inserted into the LLM’s layers, enabling efficient adaptation to specific tasks. Typically, models are fine-tuned for one task at a time, producing task-specific models or \textit{experts} that can be merged to obtain a final model supporting multiple tasks \cite{wortsman2022model,ilharco2022editing}.
A \textit{task} may correspond to, for example, (i) different domains of expertise (\eg, math versus biology Q\&A), (ii) different downstream tasks (\eg, summarization versus translation), or (iii) different optimization convergence points (\eg, results from separate training runs). We focus on LLM customization for different downstream tasks.

Model merging for multi-tasking in LLMs has gained significant attention \cite{yang2024model} due to its simplicity and ability to support multiple tasks without requiring expensive retraining or fine-tuning. By averaging or combining weights, merged models can perform tasks they were not explicitly trained on, achieving competitive performance across tasks. Techniques such as linear merging \cite{wortsman2022model,ilharco2022editing}, TIES \cite{yadav2024ties}, and DARE \cite{yu2024language} have been proposed to enhance merging efficacy.

After merging, the resulting model can switch between tasks based on user instructions. For instance, a merged model for translation and summarization can either translate or summarize input text. 
However, existing merging methods are limited to scenarios where each test example involves only one task. This limitation becomes apparent in applications where simultaneous execution of tasks is required—for example, summarizing a text while translating the summary into a different language.

We study compositional multi-tasking, a paradigm where LLMs are asked to perform multiple tasks simultaneously for a single input. As illustrated in Figure~\ref{fig:compositional}, compositional multi-tasking requires models to handle tasks such as generating a translated summary in one inference pass. While larger LLMs can directly follow instructions with an acceptable degree of success \cite{qin2024infobench}, on-device LLMs typically utilize adapters to provide acceptable output quality \cite{gunter2024apple}. We focus on the novel research direction of enabling compositional multi-tasking abilities in on-device LLMs. A simple multi-step pipeline—where one expert model performs the first task and other expert models perform the additional tasks—can achieve this (as we verify in Section~\ref{sec:experiments}), but it is inefficient, requiring several inference passes and longer processing times. 
In this paper, we study how we can achieve strong performance with a single inference pass, enabling more practical applications.

Practical real-world use cases of compositional multi-tasking for on-device LLMs abound.
For example, joint summarization and translation can be valuable when people move abroad and need to understand the overall message of a long text in the local language. Other examples include proposing suitable replies in cross-lingual scenarios, and combinations of summarization or reply with tone adjustment to suit specific contexts.
To study the novel setting of compositional multi-tasking for on-device LLMs, we developed a benchmark comprising four practical compositional tasks. Each task combines a main task (\eg, summarization or reply generation) with an auxiliary task (\eg, translation or tone adjustment). The main task determines the main functionality, while the auxiliary task modifies the output to meet additional requirements. Performance is evaluated based on the main task, with the auxiliary task altering the ground-truth texts.

Mobile devices have limited computational resources \cite{dhar2021ondevice}, but their users would benefit from on-device compositional multi-tasking functionalities.
Deploying such systems on personal devices (\eg, smartphones) introduces several \textit{desiderata}:
1) high accuracy: ensure strong task performance; 
2) low inference time: achieve compositional multi-tasking in a single inference pass over the LLM;
3) low storage requirements: minimize additional storage needs by reusing task-specialized adapters instead of creating new ones.
Therefore, training a separate adapter for each compositional task is impractical due to storage constraints on user devices. 
Nonetheless, the inefficient solutions that either perform multiple inference passes using different adapters or introduce an additional adapter to solve the compositional task will serve as reference points if we do not consider on-device restrictions. 

Further, we show existing merging strategies fail to address compositional multi-tasking in LLMs, opening the question of how we can efficiently solve the challenge.
As a solution, we propose \ours~that achieves strong performance with a single inference pass by combining already-available task-specific adapters and learning a small number of additional parameters.
This approach maintains resource efficiency while meeting the performance demands of compositional multi-tasking.

To summarize, our key contributions are: 1) We introduce the novel challenge of enabling compositional multi-tasking in on-device LLMs; 2) We develop a new benchmark with four compositional tasks to evaluate performance of different approaches; 3) We propose a novel method, \ours, which achieves high performance with minimal computational and storage overhead.

\section{Related Work}
\label{sec:related}
\keypoint{Parameter-efficient Fine-tuning (PEFT).} PEFT methods enable adapting foundational models by training only a small number of parameters, making them computationally resource-efficient \cite{han2024parameter,ding2022delta}. 
Among these, LoRA \cite{hu2021lora} has become the most widely adopted approach. LoRA introduces compact, low-rank matrices into model layers, which are trained while keeping the rest of the model frozen. This approach ensures minimal additional storage and computational requirements, making it particularly suitable for large-scale foundational models \cite{mao2025survey}.
Various extensions of LoRA have been developed to further improve its performance, for example, DoRA \cite{liu2024dora}, AdaLoRA \cite{zhang2023adalora}, Delta-LoRA \cite{zi2023delta}. Other approaches include BitFit \cite{zaken2021bitfit}, which trains the bias parameters of the foundational models.

\keypoint{Model Merging.} 
Model merging enables multi-tasking by combining multiple task-specific models into a single model. Early works \cite{wortsman2022model,ilharco2022editing} demonstrated the feasibility of linear merging, where the weights of fine-tuned expert models are combined as a weighted average. Building on this, advanced merging techniques such as TIES \cite{yadav2024ties}, DARE \cite{yu2024language}, Slerp \cite{white2016sampling}, and adaptive approaches such as LoraHub \cite{huang2023lorahub}, LM-Cocktail \cite{xiao2024lm} and Differentiable Adaptive Merging (DAM) \cite{gauthier2024merging} have been proposed. TIES resets parameters that changed too little, resolves sign conflicts, and merges only the parameters that align with the selected sign. DARE first drops part of the weight changes and then rescales the remaining ones. Slerp projects the weight changes on the sphere and then performs linear interpolation. LoraHub finds weights for linear merging via gradient-free hyperparameter optimization on a few examples. LM-Cocktail selects weights for linear merging according to loss on several examples. DAM learns column-wise scaling using backpropagation to merge multiple models.

Merging has also been explored at the level of individual adapters, allowing finer-grained combinations of model parameters \cite{tang2023parameter,zhang2023composing}. While model merging has gained traction in both NLP \cite{hammoud2024model} and computer vision \cite{yang2024model}, more advanced use cases have been developed in the vision domain, such as joint subject-style personalization with ZipLoRA \cite{shah2025ziplora} and LoRA.rar \cite{shenaj2024lora}. ZipLoRA is similar to DAM in learning column-wise scaling coefficients via backpropagation, while LoRA.rar predicts these coefficients using a pre-trained hypernetwork.
In NLP, model merging has primarily focused on standard multi-tasking, where merged models handle multiple tasks individually, with each test example addressing a single task \cite{yang2024model}. 
In contrast, we tackle the challenge of compositional simultaneous multi-tasking, where a single test example requires executing multiple tasks concurrently.

\keypoint{On-device LLMs.} 
LLMs often contain billions of parameters, necessitating significant resources, such as multiple high-end GPUs,
for inference \cite{borzunov2024distributed}. However, many valuable LLM use cases involve sensitive data stored on resource-constrained devices, making it desirable to perform computations locally and avoid transferring data to remote servers \cite{dhar2021ondevice}.
For example, users may wish to summarize private conversations or generate personalized replies while maintaining data privacy. To support such use cases, smaller LLMs have been developed for on-device deployment. These models leverage compression techniques and smaller parameter sizes to enable efficient on-device inference. Prominent examples include LLaMA 3.2 1B \cite{dubey2024llama}, Qwen2.5 1.5B \cite{qwen2,qwen2.5}, and StableLM2 1.6B \cite{bellagente2024stable}. Our work assumes that model size is the key constraint for on-device deployment, as current mobile-compatible LLMs typically range from 1–3B parameters.
Single-task LoRAs are typically stored on-device to enable an LLM to support the individual tasks, rather than relying on instruction following \cite{gunter2024apple}. On-device LLMs also have a limited context window size, making in-context learning less suitable \cite{dong2024survey}.

\section{Benchmark}
\label{sec:benchmark}

We focus on the novel problem of enabling compositional multi-tasking in on-device LLMs, which requires a suitable benchmark that includes data for both training and evaluation. To facilitate research in this domain, we develop a benchmark targeting practically valuable compositional tasks. Specifically, our benchmark includes four task combinations: summarization and reply suggestion (often referred to as ``smart reply'') as the main task $T_1$, combined with translation or tone adjustment (\ie, rewriting) as the auxiliary task $T_2$. A generic task $T$ maps an input text $x$ to an output text $y$; \ie, $T(\cdot): x \mapsto y$.
For instance, $x$ may represent a long text or part of a conversation, with the goal of producing a summary or an appropriate reply in a specific language or tone.
In general, $N$ tasks can be used, for which the compositional task would be defined as $T^C_{[N]}(x)  \eqdef T_N(\ldots T_2(T_1(x)))$, where 
\begin{equation}
T_N(\ldots T_2(T_1(x))): x \mapsto y_1 \mapsto y_2 \mapsto \ldots \mapsto y_N. 
\label{eqn:compo}
\end{equation}
Input $x$ is first processed by $T_1$ to produce $y_1$, $T_2$ subsequently transforms $y_1$ into $y_2$, \etc~ We primarily use $N=2$, but also consider $N=3$ within additional analyses.

\begin{figure*}[ht]
\vskip 0.2in
\begin{center}
\centerline{\includegraphics[trim=2cm 3
2cm 2cm 2cm, clip, width=\linewidth]{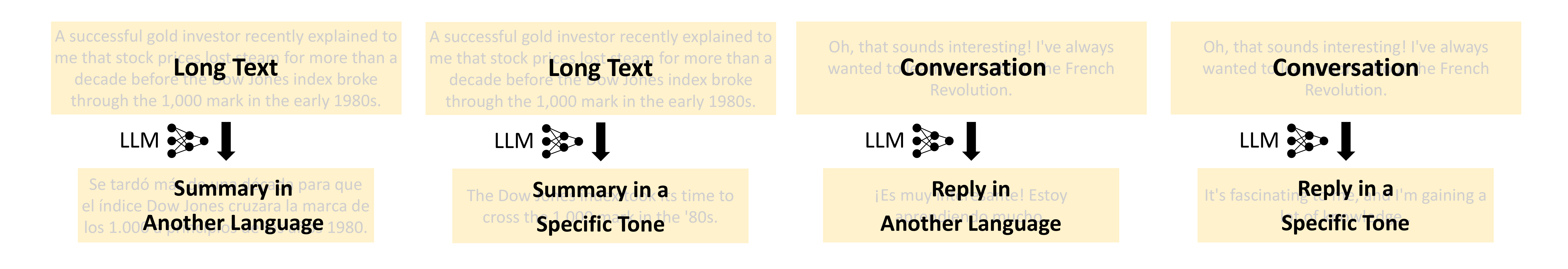}}
\caption{\textbf{Overview of the four compositional tasks in our benchmark.} The tasks include three translation settings (English to Spanish, French, and German) and four tone variations (professional, casual, witty, and neutral paraphrase), leading to fourteen sub-tasks overall.} 
\label{fig:benchmark}
\end{center}
\vskip -0.2in
\end{figure*}

Our benchmark features three translation settings (English to Spanish, French, or German) and four tone adjustments (professional, casual, witty, and paraphrase), resulting in fourteen sub-tasks in total. Existing summarization and dialogue datasets were repurposed for compositional multi-tasking using specialized models. We manually checked outputs to ensure quality and selected the most appropriate models.
Table~\ref{tab:benchmark} provides details about the tasks and the dataset splits, while Figure~\ref{fig:benchmark} illustrates the compositional tasks. 
Source datasets were selected based on their suitability and licensing terms.

\begin{table}[h]
\caption{\textbf{Dataset statistics.} Number of examples 
in our benchmark across different splits. Note that compositional tasks retain the same number of examples as their respective main tasks. For instance, summarization + translation and summarization + tone adjustment have the same number of examples as the summarization task itself, and similarly, when reply is the main task.}
\label{tab:benchmark}
\begin{center}
\resizebox{\columnwidth}{!}{
\begin{tabular}{lccc}
\toprule
\textbf{Task} & \textbf{Training} & \textbf{Validation} & \textbf{Test} \\
\midrule
Summarization & \phantom{0}12,460 & \phantom{0,}500 & 1,500 \\
Reply & 225,061 & 1,000 & 1,000 \\
\midrule
Translation & 196,026 & 4,231 & 5,571  \\
Tone adjustment & \phantom{00}2,245 & \phantom{0,}321 & \phantom{0,}642 \\
\bottomrule
\end{tabular}
}
\end{center}

\end{table}

\keypoint{Summarization + Translation.} 
We use the DialogSum dataset \cite{chen2021dialogsum}, which contains English dialogue summaries. To enable compositional multi-tasking, ground-truth summaries are translated to target languages (Spanish, French, German) using the OpusMT model \cite{tiedemann2023democratizing,TiedemannThottingal2020opus}.

\keypoint{Summarization + Tone Adjustment.} 
We change the tone of the ground-truth summaries available in DialogSum via the publicly available RedPajama-INCITE-Base 3B model \cite{weber2024redpajama} fine-tuned specifically for tone adjustment and paraphrasing \cite{utsav2023tone}. We use the corresponding prompt that was predefined for tone adjustment.

\keypoint{Reply + Translation.} 
For reply tasks, we use the Synthetic Persona Chat dataset \cite{jandaghi2023faithful} that contains a large number of dialogues. Each dialogue is converted into pairs of consecutive sentences, where the first sentence serves as the context and the second as the ground-truth reply. 
Similarly, we use the same machine translation models to translate the ground-truth outputs in target languages. To reduce computational overhead during evaluation, we sample a subset of context-reply pairs for validation and testing.

\keypoint{Reply + Tone Adjustment.} 
Tone-adjusted replies are generated from the Synthetic Persona Chat dataset using the RedPajama-INCITE-Base 3B model, \ie, the model specialized for tone adjustment (adj.) and paraphrasing.

In addition to compositional tasks, we train single task adapters on suitable datasets. Summarization and reply tasks use DialogSum and Synthetic Persona Chat, respectively. Translation tasks utilize the TED Talks dataset \cite{qi2018pretrained} (English to Spanish, French, and German). Tone adj.\ tasks are trained on the Sound Natural rephrasing dataset \cite{einolghozati2020sound}, where rephrased content is further processed using the tone adj.\ model. 

We adopt metrics commonly used in the literature to evaluate performance.
For compositional tasks based on summarization, we report the ROUGE-L (R-L) \cite{lin2004rouge} metric (\%, $\uparrow$), which measures the longest common subsequence between the generated and ground truth texts. 
Additional evaluations via ROUGE-1 and ROUGE-2 are shown in Appendix~\ref{sec:standardmetrics}.
For compositional tasks based on reply, we report the Weighted ROUGE (W-R) score \cite{zhang2021dataset} metric (\%, $\uparrow$), which is computed as: 
\begin{equation}
    \text{W-R}=\frac{\text{ROUGE-1}}{6} + \frac{\text{ROUGE-2}}{3} + \frac{\text{ROUGE-3}}{2}.
\end{equation}
Weighted ROUGE mitigates sensitivity to small sequence-length changes and has been shown \cite{zhang2021dataset} to correlate well with user click-through rates, reflecting how often the users utilize the recommended reply.

In addition, we include evaluation via LLM Judge (LLM-J) \cite{zheng2023judging}. 
LLM Judge mimics human evaluation by assessing outputs using a pretrained LLM. 
We employ Llama 3.1 70B Instruct model \cite{dubey2024llama} with carefully designed prompts (detailed in Appendix~\ref{sec:llmjudge}), assigning a binary outcome to each example.

\section{Method}
\label{sec:method}
Existing approaches that can be used for compositional multi-tasking in on-device LLMs are either inefficient or have low performance, as we show in our analysis. How can we \textit{achieve both efficiency and good performance} so that we can target on-device applications where computational resources and storage are restricted? We propose a new method that addresses this challenge.
In particular, we develop an efficient solution that achieves comparable or superior performance to inefficient baselines, such as the multi-step application of adapters or a new adapter trained specifically for each compositional task (referred to as a ``joint-expert'' adapter).

\subsection{Preliminaries}
In typical on-device scenarios, we assume access to a small LLM (\eg, 1B–3B parameters) and low-rank adapters (LoRAs) stored on devices to support tasks such as summarization or translation. 
Indeed, while prompting can be sufficient with large-scale models, LoRAs are commonly used in on-device settings to provide excellent performance in the desired tasks \cite{gunter2024apple}.
Each LoRA typically requires around 20–100 MB of storage. 
To handle compositional tasks, we aim to introduce only a negligible number of additional parameters to be stored on devices.  

LoRAs provide an efficient way to adapt LLMs by adding learnable low-rank factorized matrices ${B \in \mathbb{R}^{d\times r}}$ and $A \in \mathbb{R}^{r\times k}$ (where $r \ll \min(d, k)$) to the model's frozen weights $W_0 \in \mathbb{R}^{d\times k}$. The adjusted forward pass gives: 
\begin{equation}
    h=W_0 x+\Delta W x=W_0 x+BAx.
\end{equation}
In line with literature and real-world usage \cite{gunter2024apple}, we assume LoRAs $\{B_i, A_i\}_{i=1}^N$ for all $N$ tasks $\{T_i\}_{i=1}^N$ are already available on-device.

\subsubsection{Baseline and Merging Strategies}

We evaluate several na\"ive baselines:  
1-2) Using only the main (or auxiliary) task LoRA.  
3) In-context learning that provides an example input and output but consumes more of the context window.
4) Multi-step application of LoRAs, where an output for a task is passed as input for another task. 
5) Joint-expert LoRA trained specifically for the compositional task.  
In all cases, prompts specifying the tasks are included as part of the input.

We also consider model merging strategies, which have shown success in standard multi-tasking where test examples involve a single task at a time. 
In particular, we consider the following merging strategies applied to LoRA adapters: 
1) Linear, \ie, Model Soup \cite{wortsman2022model}; 
2) Concatenation \cite{peft}; 
3) TIES \cite{yadav2024ties}; 
4) DARE \cite{yu2024language}; 
5) Slerp \cite{white2016sampling}; 
6) LoraHub \cite{huang2023lorahub}; 
7) LM-Cocktail \cite{xiao2024lm};
8) DAM \cite{gauthier2024merging} and ZipLoRA \cite{shah2025ziplora} adapted to our use cases. 
These methods are typically designed for merging entire models rather than LoRA adapters. We adapt them to merge LoRAs and evaluate their performance on compositional tasks. Note that joint-expert LoRA, LoraHub, LM-Cocktail and DAM/ZipLoRA approaches leverage extra compositional task data.

\subsection{Our Approach: \ours}

Our method
aims to achieve strong compositional task performance while being efficient in computation and storage. 
After evaluating diverse existing merging strategies on compositional tasks, we have observed it is challenging to obtain adequate performance.
Consequently, we introduce a learnable approach that utilizes data from compositional tasks during server-side pre-training of the additional parameters.

More specifically, in order to pre-train the additional parameters to enable solving compositional tasks, we assume access to the following: 1) a base LLM, 2) task specific LoRAs $\{B_i,A_i\}_{i=1}^N$, and 3) compositional task data $\mathbb{D}^C$ obtained from the compositional task $T^C_{[N]}$ using \eqref{eqn:compo}. The data are in the form of input-output pairs $(x, y)$, where $x$ is an input such as a long text, and $y$ is the ground-truth output, such as a text summary in another language.
Hence, our approach is not data-free, unlike the various merging strategies such as TIES and  DARE, and the multi-step LoRA usage. However, the pre-training of the additional parameters is done on the server, where we can assume access to ample training data, \ie, we assume data availability is not an issue.

\begin{figure}[h]
\vskip 0.2in
\begin{center}
\centerline{\includegraphics[trim=0.0cm 0.25cm 0.0cm 0.25cm, clip, width=\linewidth]{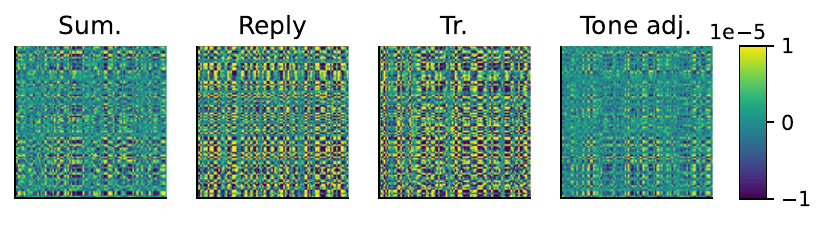}}
\caption{\textbf{LoRA weight update matrices differ across tasks,} so sharing additional parameters $P$ is likely to be suboptimal.}
\label{fig:lora_analysis}
\end{center}
\vskip -0.2in
\end{figure}

The key idea of our approach is to use merged single-task LoRAs as the initial starting point, which is then calibrated further via a relatively small number of additional parameters $P$, specific to the compositional task. As LoRA parameters vary across tasks (Figure~\ref{fig:lora_analysis}), the best performance is likely to be obtained with additional parameters $P$ that are not shared.
Our approach significantly reduces storage requirements compared to a new joint-expert adapter while enabling high performance.
The adjusted forward pass is
\begin{equation}
h=W_0x+\Delta W^c x=W_0x+f(P, \{B_i,A_i\}_{i=1}^N)x,    
\end{equation}
where $f$ represents the application of $P$ on top of the single-task LoRAs. 
The parameters $P$ are trained using compositional task data ($\mathbb{D}^C$) and a cross-entropy loss commonly used for LLM training. 
There are various options for how to utilize the parameters $P$ as described next.

\subsection{Variations of \ours}
\begin{figure}[t]
\vskip 0.2in
\begin{center}
\centerline{\includegraphics[trim=0.2cm 0.25cm 0.3cm 0.1cm, clip, width=0.9\linewidth]{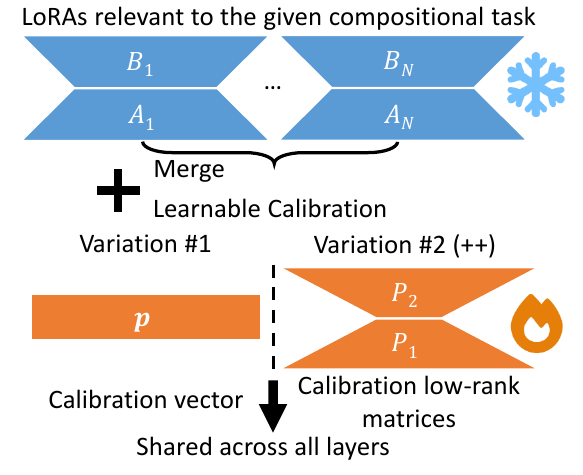}}
\caption{\textbf{Our \ours.} We add a small number of calibration parameters to correct the initial merged LoRAs. Variation \#1 uses a calibration vector of biases, while Variation \#2 (++) uses two calibration low-rank matrices.
}
\label{fig:methods}
\end{center}
\vskip -0.2in
\end{figure}

We propose two variations of \ours, providing a trade-off between model size and performance. Both variations begin with linearly merged single-task LoRAs:
\begin{equation}
B^\prime = \frac{1}{N}\sum_{i=1}^N B_i, \quad A^\prime =\frac{1}{N}\sum_{i=1}^N A_i,
\end{equation}
which are then calibrated via $f$ using parameters $P$. These parameters are shared across layers of models for efficiency, but are specific to each component, such as query or key projection, as these generally differ in size.
We provide an overview of the two variations in Figure~\ref{fig:methods}.

\keypoint{Variation \#1 (\ours).} 
The smaller variation learns a vector of column-wise biases $\boldsymbol{p} \in \mathbb{R}^d$ (a special case of $P$), applied to the LoRA update matrix $\Delta W^\prime = B^\prime A^\prime$ as 
\begin{equation}
\Delta W^c = \boldsymbol{p} \oplus B^\prime A^\prime = \boldsymbol{p} \oplus \Delta W^\prime = \sum_{i=1}^d p_i \Delta W^\prime_i,
\label{eqn:var1}
\end{equation}
where operation $\oplus$ represents element-wise column addition.
Learnable biases in $\boldsymbol{p}$ are initialized to 0.

\keypoint{Variation \#2 (\ours++).}
The larger variation introduces a calibration matrix $P$, factorized into two low-rank matrices $P_2 \in \mathbb{R}^{d\times s}$ and $P_1 \in \mathbb{R}^{s\times k}$ (where $s \ll \min(d, k)$), resulting in the following update matrix
\begin{equation}\Delta W^c = P_2 P_1 + \Delta W^\prime.
\label{eqn:var2}
\end{equation}
This effectively adds a new calibration LoRA on top of the merged single-task adapters for improved 
performance. 
The calibration parameters are initialized in the same manner as standard LoRAs and are shared across layers.

The on-device pipeline of our system involves two steps: (i) Computing the merged, calibrated LoRA $\Delta W^c$
via matrix operations \eqref{eqn:var1} or \eqref{eqn:var2}, which is extremely fast and negligible in runtime; (ii) Loading $\Delta W^c$ onto the model, which results in \textit{inference latency and throughput identical} to standard LoRAs. Our method is compatible with existing frameworks such as Android AI Core \cite{burke2023aicore} and Apple Intelligence \cite{gunter2024apple}, making it readily deployable in real-world systems.

We include an ablation study to show it is indeed beneficial to use the calibration parameters on top of the combination of single-task LoRAs -- rather than simply omitting them.

\keypoint{Sharing Calibration Parameters.}
We also consider a scenario where calibration parameters \(P\) are shared across multiple compositional tasks, \ie, the same parameters are used in all considered combinations.
In this scenario, we modify the dataset $\mathbb{D}^C$ to include examples from all four tasks in our benchmark. 
This approach further reduces storage requirements.

\section{Experiments}
\label{sec:experiments}

\subsection{Implementation Details}
We use models that are suitable for on-device settings. More specifically, we evaluate performances on our benchmark using conversational versions of LLaMA 3.2 1B, Qwen2.5 1.5B, and StableLM2 1.6B models. 
All models are multilingual and support the tested languages. 

LoRAs are applied to both the attention components (query, key, value, and output projections) and the MLP components (up, down, gate projections) \cite{fomenko2024note,Tunstall2024recipes}.
Training was conducted for one epoch on the full training set using the Adam optimizer with a learning rate of $5 \times 10^{-5}$. For LoRAs, we used a rank of $32$, a scaling factor $\alpha = 16$, and a dropout rate of $0.05$. Single-task LoRAs were trained on the datasets described in Section~\ref{sec:benchmark}.

The parameters of \ours~were trained using the Adam optimizer on a randomly selected subset of 10,000 examples from the respective compositional multi-tasking dataset, with a learning rate of $5 \times 10^{-4}$. For the \ours++ variation that incorporates factorization into two low-rank matrices, we used a rank of 4.

\subsection{Main Results}

The results of our main experiments are summarized in Table~\ref{tab:main}.
From the results, we extract the following observations: 
1) Despite including task-specific prompts, zero-shot performance was generally poor (\eg, 0.44\% LLM-J score on Sum.\ + Tr.). Similarly, in-context learning had only limited success. This suggests that on-device LLMs struggle to handle compositional tasks without fine-tuning. 
2) Using the main or auxiliary task LoRA generally improved performance compared to the zero-shot baseline (\eg, to 3.49\% in LLM-J), but remained limited in effectiveness.  
3) Both simple (\eg, TIES) and advanced (\eg, LoraHub) merging strategies performed on par with using a single LoRA, indicating their unsuitability for compositional multi-tasking.  
4) Inefficient baselines, such as the multi-step LoRA application and joint-expert LoRA, performed significantly better than the other baselines (\eg, 49.85\% and 72.92\% in LLM-J), highlighting their ability to address compositional tasks, albeit inefficiently.  
5) Our family of approaches achieved strong results (\eg, 65.15\% in LLM-J), comparable to or surpassing inefficient baselines, while being highly efficient in storage and inference passes. 
6) Overall, the more expressive \ours++~consistently achieved the best results across all benchmarks, demonstrating its robustness and adaptability. 
7) Our approaches achieve strong performance across all the different compositional tasks studied in our benchmark.

While ROUGE scores are less interpretable in isolation, the analysis of LLM judge scores provided useful insights. With \ours, tasks were successfully solved in most cases, demonstrating promising progress. Future research is needed to improve overall performance and consistency across tasks.  

\begin{table*}[ht]
\caption{\textbf{Benchmark of compositional multi-tasking.} 
Test results reported as \% ($\uparrow$) and averaged across models and languages or tones.
Our \ours~methods achieve \textit{comparable performance} to inefficient baselines while being significantly \textit{more efficient} in terms of inferences and storage.
Similarly, fast baselines, such as various merging strategies, typically fail in compositional multi-tasking. The metrics include ROUGE-L (R-L), Weighted ROUGE (W-R) and LLM judge (LLM-J) scores. The \textit{Efficient?} column captures both runtime and storage efficiency. The bottom block includes results for a version of \ours~where additional parameters are shared across all four tasks. The best three methods from the main set of results are in bold.}
\label{tab:main}
\begin{center}
\setlength{\tabcolsep}{2pt}
\resizebox{\textwidth}{!}{
\begin{tabular}{lccccccccc}
\toprule
& & \multicolumn{2}{c}{\textbf{Sum.\ + translation}} & \multicolumn{2}{c}{\textbf{Sum.\ + tone adj.}} & \multicolumn{2}{c}{\textbf{Reply + translation}} & \multicolumn{2}{c}{\textbf{Reply + tone adj.}} \\
\cmidrule(lr){3-4} \cmidrule(lr){5-6} \cmidrule(lr){7-8} \cmidrule(lr){9-10}
& \textbf{Efficient?} & R-L & LLM-J & R-L & LLM-J & W-R & LLM-J & W-R & LLM-J \\
\midrule
Zero-shot & \cmark & \phantom{0}7.49 & \phantom{0}0.44 & 13.93 & \phantom{0}6.52 & \phantom{0}2.03 & \phantom{0}4.11 & \phantom{0}3.53 & 33.66 \\
Main-task LoRA & \cmark & 13.39 & \phantom{0}3.49 & 16.43 & \phantom{0}4.18 & \phantom{0}1.25 & \phantom{0}7.17 & \phantom{0}9.12 & 36.25 \\
Auxiliary-task LoRA & \cmark & 13.99 & \phantom{0}0.30 & 14.33 & \phantom{0}5.81 & \phantom{0}4.03 & \phantom{0}4.73 & \phantom{0}4.59 & 36.68 \\
In-context learning & \cmark & 14.46 & 10.95 & 16.96 & 24.12 & \phantom{0}3.93 & \phantom{0}8.72 & \phantom{0}4.66 & 46.23 \\
\hdashline
Linear merge & \cmark & 14.27 & \phantom{0}0.33 & 15.26 & \phantom{0}2.74 & \phantom{0}3.94 & 12.81 & \phantom{0}7.72 & 41.93 \\
Concat merge & \cmark & 14.39 & \phantom{0}0.34 & 15.27 & \phantom{0}2.76 & \phantom{0}3.97 & 13.10 & \phantom{0}7.65 & 41.77 \\
TIES merge & \cmark  & 12.25 & \phantom{0}0.81 & 14.95 & \phantom{0}6.06 & \phantom{0}3.53 & \phantom{0}8.30 & \phantom{0}6.47 & \textbf{47.87} \\
DARE merge & \cmark & \phantom{0}9.27 & \phantom{0}0.68 & 14.51 & \phantom{0}6.34 & \phantom{0}2.95 & \phantom{0}8.57 & \phantom{0}4.56 & 41.76 \\
Slerp merge & \cmark & 13.96 & \phantom{0}0.54 & 14.87 & \phantom{0}4.87 & \phantom{0}3.73 & \phantom{0}9.04 & \phantom{0}6.57 & 46.97 \\
LoraHub merge & \cmark & 13.78 & \phantom{0}1.59 & 16.13 & \phantom{0}3.03 & \phantom{0}3.26 & 12.69 & \phantom{0}8.69 & 39.07 \\
LM-Cocktail merge & \cmark & 13.62 & \phantom{0}0.70 & 14.88 & \phantom{0}5.70 & \phantom{0}3.24 & \phantom{0}7.93 & \phantom{0}6.67 & \textbf{48.76} \\
DAM/ZipLoRA merge & \cmark & 16.19 & 17.16 & 15.68 & \phantom{0}2.78 & \phantom{0}4.62 & 31.88 & \phantom{0}8.00 & 43.19 \\
\hdashline
Multi-step LoRA usage & \xmark & 21.25 & \textbf{72.92} & \textbf{20.23} & \textbf{34.32} & \textbf{10.04} & \textbf{69.83} & \phantom{0}8.09 & 45.78 \\
Joint-expert LoRA & \xmark  & \textbf{21.36} & 49.85 & 19.08 & 16.14 & \textbf{14.99} & \textbf{65.73} & \textbf{14.33} & \textbf{47.06} \\
\hdashline
\ours & \cmark & \textbf{25.40} & \textbf{59.23} & \textbf{24.58} & \textbf{28.89} & \phantom{0}8.85 & 57.46 & \textbf{10.86} & 44.99 \\
\ours++ & \cmark & \textbf{28.64} & \textbf{65.15} & \textbf{26.96} & \textbf{34.34} & \textbf{12.14} & \textbf{63.81} & \textbf{13.54} & 45.40 \\
\midrule
Shared \ours & \cmark & 17.04 & 29.91 & 20.98 & 16.86 & \phantom{0}6.63 & 49.23 & \phantom{0}9.23 & 39.61 \\
Shared \ours++ & \cmark & 19.23 & 32.61 & 22.68 & 15.54 & \phantom{0}9.99 & 56.74 & 11.06 & 42.33 \\
\bottomrule
\end{tabular}
}
\end{center}
\end{table*}

\subsection{Analyses}
\label{sec:analyses}

We have performed various further analyses, including ones about sharing of parameters across tasks, efficiency, benefit of using existing adapters and analysis of changes in the update matrices.

\keypoint{Sharing Parameters Across Tasks.}
To improve efficiency further, we investigate if \ours~parameters can be shared across tasks. Results from the bottom block of Table~\ref{tab:main} show that sharing parameters leads to a slight performance decrease compared to using task-specific parameters.
Nevertheless, shared parameters still outperform most baseline approaches, demonstrating their potential as a more storage-efficient solution.

\keypoint{Efficiency Analysis.}
We compare the efficiency of our solutions against the two well-performing but inefficient baselines in Table~\ref{tab:efficiency}. 
Our solutions require only a minimal number of additional parameters, amounting to approximately 0.08–0.56\% of the parameters of a joint-expert LoRA. The resulting additional storage on disk is less than 0.5 MB, making our solutions suitable for on-device deployment.

\begin{table}[t]
\caption{\textbf{Efficiency of well-performing approaches.} Our methods require only 0.08–0.56\% of additional parameters/storage, depending on the variation (averaged across models). Baselines not reported here, such as \textit{Main-task LoRA} and \textit{Linear Merge}, are efficient (\ie, single inference pass, zero additional parameters/storage) but have significantly lower performance.}
\label{tab:efficiency}
\setlength{\tabcolsep}{2pt}
\begin{center}
\resizebox{\columnwidth}{!}{
\begin{tabular}{lccc}
\toprule
\textbf{Method} & \multicolumn{1}{p{2cm}}{\centering \textbf{\# of}  \\ \textbf{Inferences}} &  \multicolumn{1}{p{2cm}}{\centering \textbf{Additional} \\ \textbf{Parameters}} & \multicolumn{1}{p{2cm}}{\centering \textbf{Additional} \\ \textbf{Storage}} \\
\midrule
Multi-step LoRAs & $2\times$ & 0 & \phantom{0}0.00MB \\
Joint-expert LoRA & $1\times$ & \phantom{0}30M & 57.10MB \\
\hdashline
\ours & $1\times$ & \phantom{0}23K & \phantom{0}0.05MB \\
\ours++ & $1\times$ & 166K & \phantom{0}0.32MB \\
\bottomrule
\end{tabular}
}
\end{center}
\end{table}

\keypoint{Benefit of Using Existing Adapters.}
To assess the impact of using existing adapters, we evaluate performance when training only the \ours~parameters without leveraging pre-existing adapters. Results in Table~\ref{tab:ablation} confirm that starting with existing adapters and calibrating them using learnable parameters (biases or low-rank matrices) improves performance. 
Without calibration parameters, the approach becomes linear merge that has weak performance, highlighting the critical role of calibration for achieving strong results. The best performance is obtained when the merged adapters are used as a starting point and subsequently refined with learnable calibration parameters.  

\begin{table*}[h!]
\caption{\textbf{Effect of pre-existing adapters.} Leveraging pre-existing adapters enhances performance in compositional multi-tasking for both variations of \ours~(LC). Results 
are 
averaged across different models and languages or tones. The top block shows results with separate learned parameters for each compositional task, while the bottom block shows results with shared parameters across tasks.}
\label{tab:ablation}
\begin{center}
\resizebox{0.88\textwidth}{!}{
\begin{tabular}{clcccccccc}
\toprule
& & \multicolumn{2}{c}{\textbf{Sum.\ + translation}} & \multicolumn{2}{c}{\textbf{Sum.\ + tone adj.}} & \multicolumn{2}{c}{\textbf{Reply + translation}} & \multicolumn{2}{c}{\textbf{Reply + tone adj.}} \\
\cmidrule(lr){3-4} \cmidrule(lr){5-6} \cmidrule(lr){7-8} \cmidrule(lr){9-10}
& & R-L & LLM-J & R-L & LLM-J & W-R & LLM-J & W-R & LLM-J \\
\midrule
\multirow{4}{*}{\rotatebox{90}{Separate}} & LC & 25.40 & 59.23 & 24.58 & 28.89 & \phantom{0}8.85 & 57.46 & 10.86 & 44.99 \\
& LC w/o LoRAs & 25.21 & 57.96 & 23.62 & 26.81 & \phantom{0}7.36 & 53.18 & \phantom{0}9.88 & 40.13 \\
& LC++ & 28.64 & 65.15 & 26.96 & 34.34 & 12.14 & 63.81 & 13.54 & 45.40 \\
& LC++ w/o LoRAs & 28.45 & 63.20 & 26.96 & 34.73 & 11.76 & 61.82 & 13.13 & 44.62 \\
\hline
\multirow{4}{*}{\rotatebox{90}{Shared}} & LC & 17.04 & 29.91 & 20.98 & 16.86 & \phantom{0}6.63 & 49.23 & \phantom{0}9.23 & 39.61 \\
& LC w/o LoRAs & 15.36 & 24.31 & 19.24 & 11.81 & \phantom{0}5.83 & 42.23 & \phantom{0}8.44 & 39.12 \\
& LC++ & 19.23 & 32.61 & 22.68 & 15.54 & \phantom{0}9.99 & 56.74 & 11.06 & 42.33 \\
& LC++ w/o LoRAs & 17.70 & 35.66 & 21.53 & 16.28 & \phantom{0}9.51 & 54.79 & 10.90 & 42.17 \\
\hline
\end{tabular}
}
\end{center}
\end{table*}

\begin{figure}[!h]
\vskip 0.2in
\begin{center}
\begin{subfigure}[h]{\columnwidth}
\centerline{\includegraphics[width=\linewidth]{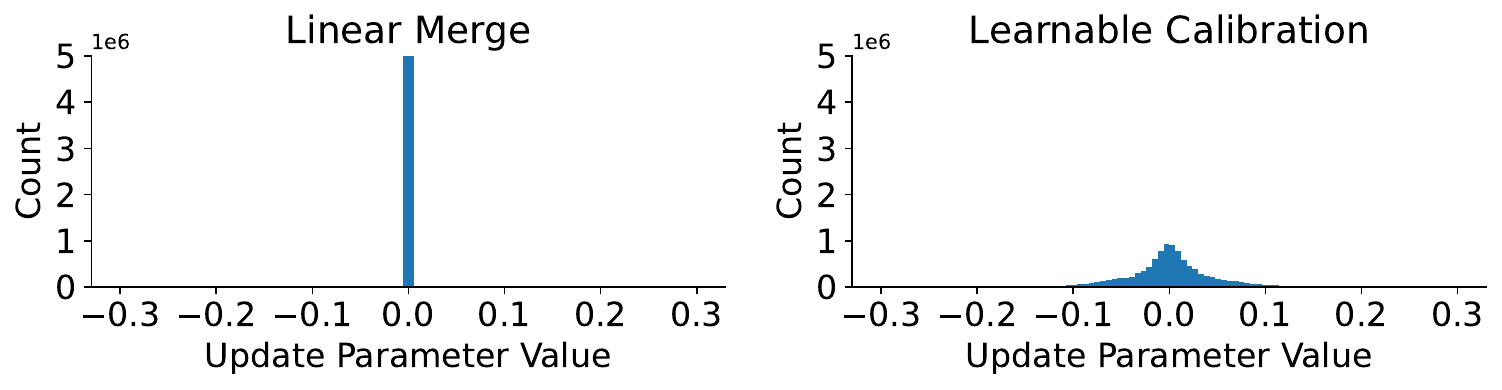}}
\centerline{\includegraphics[width=\linewidth]{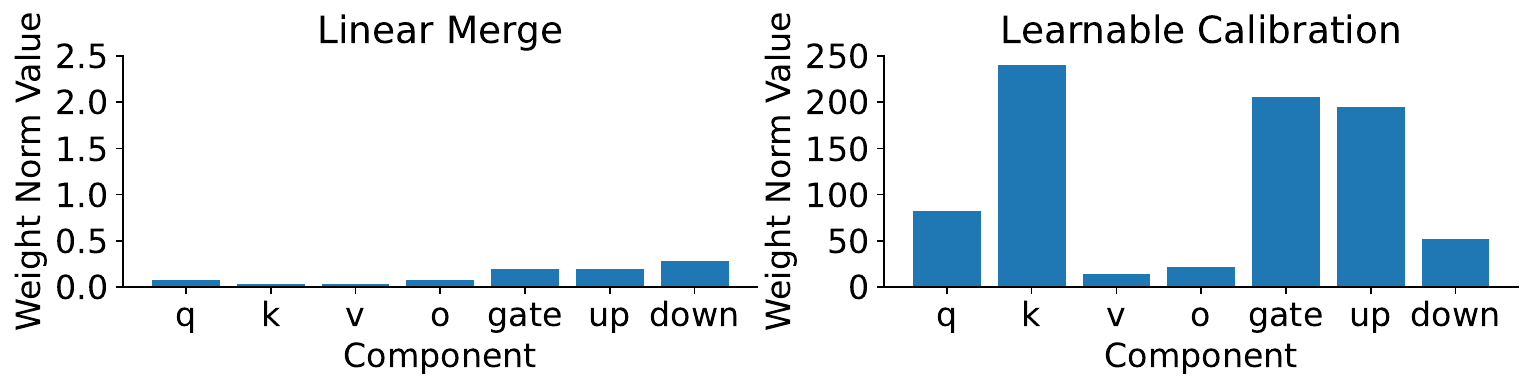}}
\caption{Sum. + translation with Qwen2.5 1.5B model.}
\end{subfigure}
\begin{subfigure}[h]{\columnwidth}
\centerline{\includegraphics[width=\linewidth]{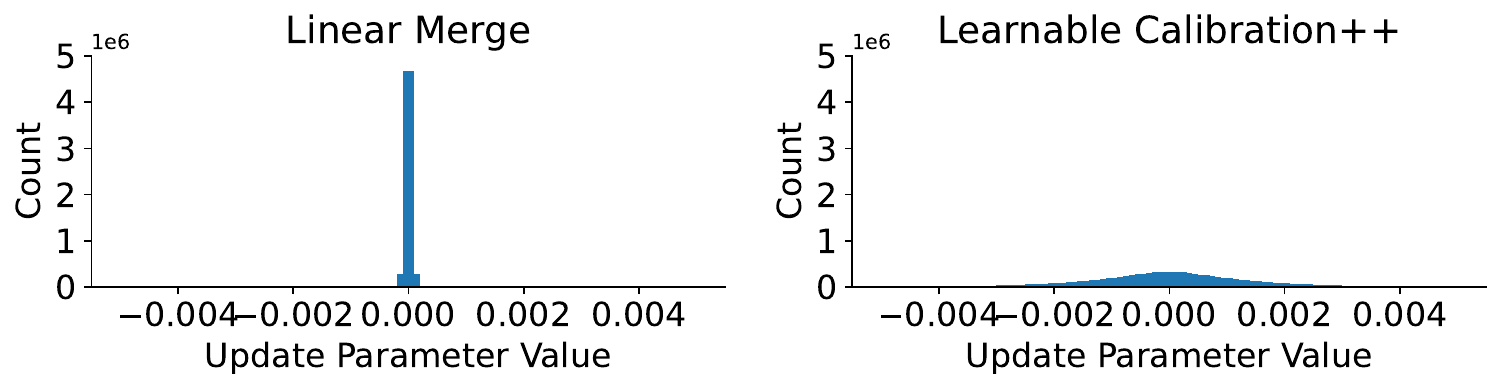}}
\centerline{\includegraphics[width=\linewidth]{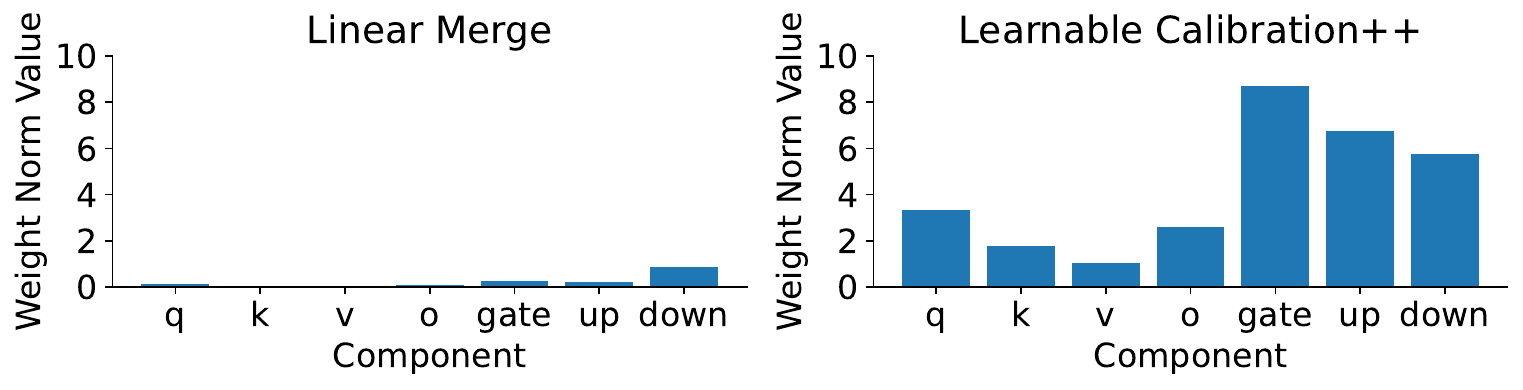}}
\caption{Reply + tone adj. with Llama 3.2 1B.}
\end{subfigure}
\caption{\textbf{Changes in the update matrix.} \ours~makes the overall LoRA update matrix significantly more diverse to handle additional tasks. The weight norms also increase substantially across different components, reflecting the added complexity required for compositional multi-tasking.
}
\label{fig:behaviour}
\end{center}
\vskip -0.3in
\end{figure}

\keypoint{Analysis of Changes in the Update Matrices.}
To understand how our methods address compositional tasks, we analyze the changes in the update matrices compared to simple linear merging. We examine the distribution of values of the parameters via histograms and evaluate weight norms for two scenarios: sum. + tr. and reply + tone adj. (Figure~\ref{fig:behaviour}).  
We observe that the addition of calibration parameters significantly increases the diversity of the LoRA update matrix, enabling the handling of auxiliary tasks. Also, the weight norms of the update matrices grow substantially, reflecting the additional complexity required to address compositional tasks.
These patterns are consistent across both scenarios, illustrating how \ours~calibrates the adapters to accommodate multiple tasks effectively.

\keypoint{Further Analyses.}
Notably, we show \ours~1) works well for models of various sizes such as from 0.5B to 3B (Appendix~\ref{sec:scaling}), 2) is able to handle domain shift (Appendix~\ref{sec:ood}), and 3) obtains strong performance also for three-way compositional tasks (Appendix~\ref{sec:threeway}). We also perform qualitative analysis in Appendix~\ref{sec:qualitative}. The analysis shows there are essentially two behavioral groups. The first group typically does not perform one of the tasks correctly and includes approaches such as zero-shot and existing merging strategies. The second group succeeds in performing both tasks and includes the two inefficient baselines as well as our \ours~solutions.

\section{Conclusion}
\label{sec:conclusion}
We introduced the practically valuable problem of compositional multi-tasking for LLMs in on-device settings, where computational and storage resources are constrained.
To facilitate research in this area, we developed a comprehensive benchmark comprising diverse practical compositional tasks. Our evaluation demonstrates that existing methods either lack efficiency or fail to achieve adequate performance, highlighting the need for new approaches.  
As a solution, we proposed \ours, a family of methods that leverage pre-existing adapters on a device and calibrate them with a minimal number of additional parameters, achieving strong performance while being efficient in both storage and computation.

\section*{Limitations}
Our evaluation focused on on-device-sized LLMs as that is the envisioned use case. One could also perform evaluation with significantly larger models, but this would require significant compute resources. We studied combinations of two and three tasks, as these are the most practically relevant, but it could be possible to evaluate with a larger number of tasks too if our benchmark was extended.

\section*{Ethical Considerations}
Our research lies in the domain of text-generative AI, a field with significant ethical and societal implications. The proposed use cases for compositional multi-tasking, such as summarization and translation, have the potential to benefit users across diverse applications. However, this versatility also introduces risks, as the technology could be misused to achieve undesirable goals.  

While LLMs typically incorporate built-in safeguarding mechanisms, these safeguards may be weakened when models are merged \cite{hammoud2024model} or adapted for compositional multi-tasking. It is crucial to ensure that safeguards remain effective in these scenarios before deploying solutions.  

Potential future research could explore the interplay between compositional multi-tasking and safeguarding mechanisms, with a focus on developing robust, high-performing safeguards for the proposed setup. 
Addressing these challenges will be essential for the responsible advancement and deployment of compositional multi-tasking capabilities. 

\bibliography{custom}

\newpage
\appendix

\renewcommand{\thefigure}{A\arabic{figure}}
\renewcommand{\theHfigure}{A\arabic{figure}}
\renewcommand{\theequation}{A\arabic{equation}}
\renewcommand{\theHequation}{A\arabic{equation}}
\renewcommand{\thetable}{A\arabic{table}}
\renewcommand{\theHtable}{A\arabic{table}}

\setcounter{equation}{0}
\setcounter{figure}{0}
\setcounter{table}{0}

\clearpage

\section{LLM Judge Evaluation}
\label{sec:llmjudge}

We conducted an evaluation using the LLaMA 3.1 70B instruction-tuned model \cite{dubey2024llama} as an LLM Judge. The evaluation prompts used for the main scenarios are provided in Figures~\ref{fig:llm-judge}, \ref{fig:llm-judge2}. The LLM Judge prompt used for evaluating compositional multi-tasking with three tasks is shown in Figure~\ref{fig:llm-judge-three}.

Given the large number of experiments conducted, LLM Judge evaluations are relatively expensive to run locally, particularly for researchers operating with limited computational resources. As such, we recommend focusing on standard metrics such as ROUGE-L and Weighted ROUGE when developing and testing new approaches using our benchmark.

Our LLM judge results provide an additional layer of interpretation, offering a practical (albeit somewhat noisy) perspective on the scores derived from standard metrics. These results help bridge the gap between quantitative evaluation and qualitative performance assessment.

\begin{figure*}[t]
    \centering
\begin{tcbraster}[raster columns=2, raster halign=center, raster equal height=all]
\begin{tcolorbox}[width=.475\textwidth, nobeforeafter,colback=green!5!white, colframe=green!75!black, title=\textbf{Prompt for Cross-Lingual Summarization}]

You are in a setting where you want to summarize a dialogue in another language. The dialogue is in English, but the summary should be in \texttt{\{language\}}. Your task is to judge the quality of the summary and if it is in \texttt{\{language\}}.

The summary should not take the form of a reply to the dialogue.

I give you an example ground-truth summary that is in \texttt{\{language\}} and would be considered a good summary of the dialogue\\

Dialogue:  \texttt{\{dialogue\}}\\

Ground truth summary: \texttt{\{ground truth summary\}}\\

Proposed summary: \texttt{\{proposed summary\}}\\

Does the proposed summary summarize the dialogue and is it in \texttt{\{language\}}? Please respond with 0 or 1. Only output 0 or 1 and nothing else.

\end{tcolorbox}
\begin{tcolorbox}[width=.475\textwidth,colback=green!5!white, colframe=green!75!black, title=\textbf{Prompt for Cross-Tone Summarization}]

You are in a setting where you want to summarize a dialogue in a \texttt{\{tone\}} tone. Your task is to judge the quality of the summary and if it uses a \texttt{\{tone\}} tone.

The summary should not take the form of a reply to the dialogue.

I give you an example ground-truth summary that follows a \texttt{\{tone\}} tone and would be considered a good summary of the dialogue.\\

Dialogue: \texttt{\{dialogue\}}\\

Ground-truth summary: \texttt{\{ground-truth summary\}}\\

Proposed summary: \texttt{\{proposed summary\}}\\

Does the proposed summary summarize the dialogue and is it written in a \texttt{\{tone\}} tone? Please respond with 0 or 1. Only output 0 or 1 and nothing else.

\end{tcolorbox}
\end{tcbraster}
    \caption{\textbf{LLM Judge prompts for summarization-based compositional tasks.} Prompts used by LLaMA 3.1 70B LLM Judge to evaluate compositional tasks with a given target language or tone.}
    \label{fig:llm-judge}
\end{figure*}

\begin{figure*}[t]
    \centering
\begin{tcbraster}[raster columns=2, raster halign=center, raster equal height=all]
\begin{tcolorbox}[width=.475\textwidth,colback=green!5!white, colframe=green!75!black, title=\textbf{Prompt for Cross-Lingual Reply Suggestion}]

You are in a setting where you receive a message in English and you reply in \texttt{\{language\}}. Your task is to judge the usefulness of the proposed reply.

I give you an example ground-truth response in \texttt{\{language\}} that would be considered a good reply for the message in English.

The proposed reply should not directly translate or paraphrase the message you have received, it should respond to it.\\

Message in English: \texttt{\{message\}}\\

Ground-truth reply in \texttt{\{language\}}: \texttt{\{ground-truth reply\}}\\

Proposed reply: \texttt{\{proposed reply\}}\\

Is the proposed reply useful and is it in \texttt{\{language\}}? Please respond with 0 or 1. Only output 0 or 1 and nothing else.

\end{tcolorbox}
\begin{tcolorbox}[width=.475\textwidth,colback=green!5!white, colframe=green!75!black, title=\textbf{Prompt for Cross-Tone Reply Suggestion}]

You are in a setting where you receive a message and you want to reply in a \texttt{\{tone\}} tone. Your task is to judge the usefulness of the proposed reply.

I give you an example ground-truth response that follows a \texttt{\{tone\}} tone and would be considered a good reply for the message.\\

Message: \texttt{\{message\}}\\

Ground-truth reply: \texttt{\{ground-truth reply\}}\\

Proposed reply: \texttt{\{proposed reply\}}\\

Is the proposed reply relevant and is it written in a \texttt{\{tone\}} tone? Please respond with 0 or 1. Only output 0 or 1 and nothing else.
\\
\\
\\
\\
\end{tcolorbox}
\end{tcbraster}
    \caption{\textbf{LLM Judge prompts for reply-based compositional tasks.} Prompts used by LLaMA 3.1 70B LLM Judge to evaluate compositional tasks with a given target language or tone.}
    \label{fig:llm-judge2}
\end{figure*}

\begin{figure*}[t]
    \centering
\begin{tcolorbox}[width=\textwidth, nobeforeafter,colback=green!5!white, colframe=green!75!black, title=\textbf{Prompt for Cross-Lingual Tone-Adjusted Summarization}]

You are in a setting where you want to summarize a dialogue in a \texttt{\{tone\}} tone in another language. The dialogue is in English, but the summary should be in a \texttt{\{tone\}} tone in \texttt{\{language\}}. Your task is to judge the quality of the summary, if it uses a \texttt{\{tone\}} tone and if it is in \texttt{\{language\}}.

The summary should not take the form of a reply to the dialogue.

I give you an example ground-truth summary that follows a \texttt{\{tone\}} tone and is in \texttt{\{language\}}. The example would be considered a good summary of the dialogue.\\

Dialogue: \texttt{\{dialogue\}}\\

Ground-truth summary: \texttt{\{ground-truth summary\}}\\

Proposed summary: \texttt{\{proposed summary\}}\\

Does the proposed summary summarize the dialogue and is it written in a \texttt{\{tone\}} tone in \texttt{\{language\}}? Please respond with 0 or 1. Only output 0 or 1 and nothing else.

\end{tcolorbox}
    \caption{\textbf{LLM Judge prompt for three-way compositional tasks.} Prompt used by LLaMA 3.1 70B LLM Judge to evaluate three compositional tasks with a given target language and tone.}
    \label{fig:llm-judge-three}
\end{figure*}

\section{Additional Details}
\label{sec:additional-details}

\subsection{Task Prompts}
For all compositional tasks, we include a prompt explicitly specifying the desired combination of tasks. The prompts used are as follows:

\begin{itemize}
    \item \textbf{Summarization + translation}: \textit{Summarize the following text and translate it from English to} \texttt{\{language\}}.
    \item \textbf{Summarization + tone adjustment}: \textit{Summarize the following text and change its tone to} \texttt{\{tone\}}
    \item \textbf{Reply + translation}: \textit{Suggest a reply for the following text and translate it from English to} \texttt{\{language\}}
    \item \textbf{Reply + tone adjustment}: \textit{Suggest a reply for the following text and change its tone to} \texttt{\{tone\}}
\end{itemize}

In the case of in-context learning, we also provide one example for the task, showing an example input and output. One example was used because on-device LLMs have a very limited context window, and so in-context learning is generally not a suitable strategy in these settings.

\subsection{Hyperparameter Selection}

Hyperparameters were selected using validation data. For common merging strategies, such as linear merging and TIES, we tested a wide range of hyperparameters on a representative compositional multi-tasking scenario. However, we observed that performance was consistently similar across different configurations and did not surpass the performance of using either the main or auxiliary-task LoRA paired with prompting.

As a result, we chose hyperparameters that are broadly applicable across diverse tasks: weights of 0.5 (linear, concat, TIES, DARE merges) and density of 0.5 (TIES, DARE merges).
These values are intended to represent a reasonable balance, ensuring fair comparisons across different methods.

\section{Extended Evaluation with Standard Metrics}
\label{sec:standardmetrics}
We provide an extended evaluation using all standard metrics, including ROUGE-1 (R-1), ROUGE-2 (R-2), ROUGE-L (R-L), and Weighted ROUGE (W-R). 
The detailed results are presented in Table~\ref{tab:main_rouge} and Table~\ref{tab:ablation_rouge}. 
These results are consistent with those reported in the main text, further validating the effectiveness and robustness of our proposed methods.

\begin{table*}[h!]
\caption{\textbf{Evaluation of compositional multi-tasking across different scenarios.} Test results reported as percentages (\%, \(\uparrow\)) and averaged across different models and languages or tones. Our family of methods achieves comparable performance to inefficient baselines while being significantly more efficient in terms of inferences and storage. Similarly, fast baselines, such as various merging strategies, generally fail in compositional multi-tasking. The bottom block includes results for a version where the additional parameters of \ours~are shared across all four tasks.}
\label{tab:main_rouge}
\begin{center}
\resizebox{\textwidth}{!}{
\setlength{\tabcolsep}{3pt}
\begin{tabular}{lcccccccc}
\toprule
& \multicolumn{3}{c}{\textbf{Sum.\ + translation}} & \multicolumn{3}{c}{\textbf{Sum.\ + tone adj.}} & \multicolumn{1}{c}{\textbf{Reply + translation}} & \multicolumn{1}{c}{\textbf{Reply + tone adj.}} \\
\cmidrule(lr){2-4} \cmidrule(lr){5-7} \cmidrule(lr){8-8} \cmidrule(lr){9-9}
& \textbf{R-1} & \textbf{R-2} & \textbf{R-L} & \textbf{R-1} & \textbf{R-2} & \textbf{R-L} & \textbf{W-R} & \textbf{W-R} \\
\midrule
Zero-shot & \phantom{0}8.68 & \phantom{0}1.45 & \phantom{0}7.49 & 18.84 & \phantom{0}3.42 & 13.93 & \phantom{0}2.03 & \phantom{0}3.53 \\
Main-task LoRA & 16.33 & \phantom{0}2.73 & 13.39 & 21.53 & \phantom{0}4.43 & 16.43 & \phantom{0}1.25 & \phantom{0}9.12 \\
Auxiliary-task LoRA & 16.77 & \phantom{0}2.98 & 13.99 & 18.61 & \phantom{0}3.71 & 14.33 & \phantom{0}4.03 & \phantom{0}4.59 \\
In-context learning & 18.05 & \phantom{0}3.76 & 14.46 & 23.53 & \phantom{0}5.68 & 16.96 & \phantom{0}3.93 & \phantom{0}4.66 \\
\hdashline
Linear merge & 16.90 & \phantom{0}3.02 & 14.27 & 19.84 & \phantom{0}3.93 & 15.26 & \phantom{0}3.94 & \phantom{0}7.72 \\
Concat merge & 17.04 & \phantom{0}3.08 & 14.39 & 19.87 & \phantom{0}3.94 & 15.27 & \phantom{0}3.97 & \phantom{0}7.65 \\
TIES merge & 14.35 & \phantom{0}2.65 & 12.25 & 20.03 & \phantom{0}3.88 & 14.95 & \phantom{0}3.53 & \phantom{0}6.47 \\
DARE merge & 10.78 & \phantom{0}1.96 & \phantom{0}9.27 & 19.50 & \phantom{0}3.69 & 14.51 & \phantom{0}2.95 & \phantom{0}4.56 \\
Slerp merge & 16.40 & \phantom{0}2.97 & 13.96 & 19.68 & \phantom{0}3.81 & 14.87 & \phantom{0}3.73 & \phantom{0}6.57 \\
LoraHub merge & 16.51 & \phantom{0}2.80 & 13.78 & 20.94 & \phantom{0}4.32 & 16.13 & \phantom{0}3.26 & \phantom{0}8.69 \\
LM-Cocktail merge & 15.97 & \phantom{0}2.88 & 13.62 & 19.78 & \phantom{0}3.81 & 14.88 & \phantom{0}3.24 & \phantom{0}6.67 \\
DAM/ZipLoRA merge & 20.38 & \phantom{0}4.38 & 16.19 & 20.36 & \phantom{0}4.14 & 15.68 & \phantom{0}4.62 & \phantom{0}8.00 \\
\hdashline
Multi-step LoRA usage & 27.20 & \phantom{0}7.92 & 21.25 & 27.19 & \phantom{0}6.85 & 20.23 & 10.04 & \phantom{0}8.09 \\
Joint-expert LoRA & 26.77 & \phantom{0}7.05 & 21.36 & 24.98 & \phantom{0}6.38 & 19.08 & 14.99 & 14.33 \\
\hdashline
\ours & 31.83 & \phantom{0}9.53 & 25.40 & 33.14 & \phantom{0}9.48 & 24.58 & \phantom{0}8.85 & 10.86 \\
\ours++ & 35.57 & 12.11 & 28.64 & 36.09 & 10.95 & 26.96 & 12.14 & 13.54 \\
\midrule
Shared \ours & 21.15 & \phantom{0}5.30 & 17.04 & 28.00 & \phantom{0}7.72 & 20.98 & \phantom{0}6.63 & \phantom{0}9.23 \\
Shared \ours++ & 23.91 & \phantom{0}6.25 & 19.23 & 29.89 & \phantom{0}8.38 & 22.68 & \phantom{0}9.99 & 11.06 \\
\bottomrule
\end{tabular}
}
\end{center}
\end{table*}

\begin{table*}[h!]
\caption{\textbf{Ablation on usefulness of single-task LoRAs.} Leveraging existing adapters enhances the performance of \ours~(LC) in compositional multi-tasking. Test scores (\%, $\uparrow$) are reported, averaged across various models and languages or tones. The top block utilizes separate parameters for each compositional task, while the bottom block employs a single set of shared parameters.}
\label{tab:ablation_rouge}
\begin{center}
\resizebox{\textwidth}{!}{
\setlength{\tabcolsep}{4pt}
\begin{tabular}{clcccccccc}
\toprule
&& \multicolumn{3}{c}{\textbf{Sum.\ + translation}} & \multicolumn{3}{c}{\textbf{Sum.\ + tone adj.}} & \multicolumn{1}{c}{\textbf{Reply + translation}} & \multicolumn{1}{c}{\textbf{Reply + tone adj.}} \\
\cmidrule(lr){3-5} \cmidrule(lr){6-8} \cmidrule(lr){9-9} \cmidrule(lr){10-10}
&& \textbf{R-1} & \textbf{R-2} & \textbf{R-L} & \textbf{R-1} & \textbf{R-2} & \textbf{R-L} & \textbf{W-R} & \textbf{W-R} \\
\midrule
\multirow{4}{*}{\rotatebox{90}{Separate}} & LC & 31.83 & \phantom{0}9.53 & 25.40 & 33.14 & \phantom{0}9.48 & 24.58 & \phantom{0}8.85 & 10.86 \\
& LC w/o LoRAs & 31.53 & \phantom{0}9.36 & 25.21 & 32.02 & \phantom{0}8.95 & 23.62 & \phantom{0}7.36 & \phantom{0}9.88 \\
& LC++ & 35.57 & 12.11 & 28.64 & 36.09 & 10.95 & 26.96 & 12.14 & 13.54 \\
& LC++ w/o LoRAs & 35.23 & 11.94 & 28.45 & 36.08 & 10.95 & 26.96 & 11.76 & 13.13 \\
\midrule
\multirow{4}{*}{\rotatebox{90}{Shared}} & LC & 21.15 & \phantom{0}5.30 & 17.04 & 28.00 & \phantom{0}7.72 & 20.98 & \phantom{0}6.63 & \phantom{0}9.23 \\
& LC w/o LoRAs & 19.05 & \phantom{0}4.73 & 15.36 & 25.53 & \phantom{0}6.76 & 19.24 & \phantom{0}5.83 & \phantom{0}8.44 \\
& LC++ & 23.91 & \phantom{0}6.25 & 19.23 & 29.89 & \phantom{0}8.38 & 22.68 & \phantom{0}9.99 & 11.06 \\
& LC++ w/o LoRAs & 21.91 & \phantom{0}5.73 & 17.70 & 28.31 & \phantom{0}7.83 & 21.53 & \phantom{0}9.51 & 10.90 \\
\bottomrule
\end{tabular}
}
\end{center}
\end{table*}

\clearpage
\clearpage

\section{Scaling Analysis}
\label{sec:scaling}
\begin{figure*}[ht]
\vskip 0.2in
\begin{center}
\centerline{\includegraphics[width=0.96\linewidth]{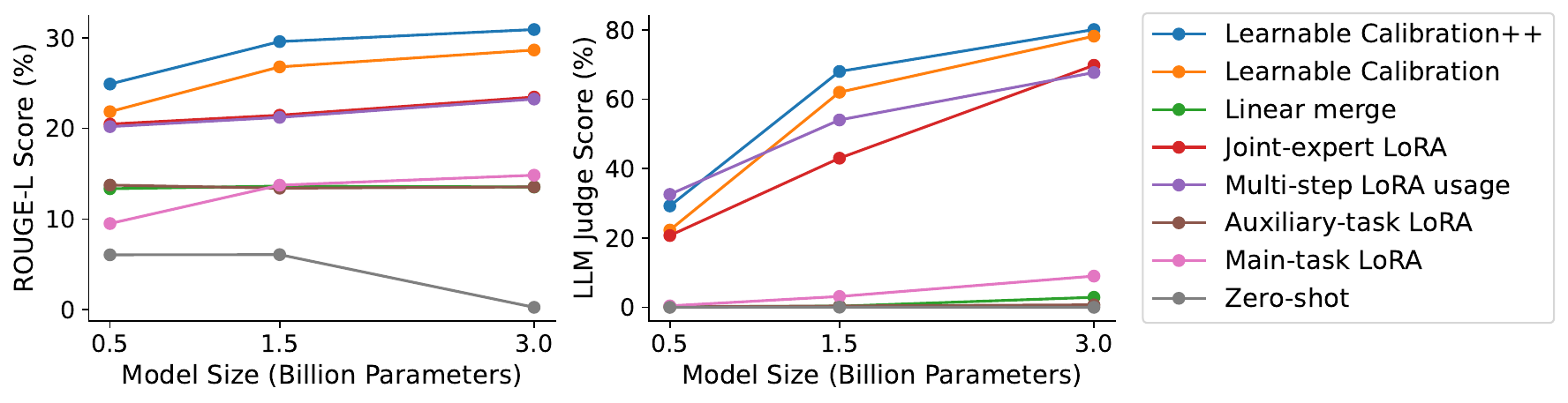}}
\caption{\textbf{Evaluation across model sizes.} Evaluation of compositional multi-tasking on the cross-lingual summarization task for the Qwen2.5 model across different sizes: 0.5B, 1.5B, and 3B. Test results for ROUGE-L and LLM Judge score (\%, \(\uparrow\)) are averaged across languages. Our solutions consistently demonstrate strong performance across all model sizes, with performance improving as model size increases.}
\label{fig:scaling}
\end{center}
\vskip -0.2in
\end{figure*}

LLMs designed for mobile device deployment are available in various sizes, including newer models with up to 3B parameters \cite{gunter2024apple,carreira2023revolutionizing}. We analyze the scalability of our solution across different model sizes in Figure~\ref{fig:scaling}.
For this analysis, we use the Qwen2.5 model, which is available in three sizes suitable for on-device deployment: 0.5B, 1.5B, and 3B parameters. The results demonstrate that our solutions consistently perform well across all model sizes. Additionally, as expected, larger models typically achieve stronger performance.

\section{Out-of-Domain Generalization}
\label{sec:ood}
We evaluate the generalization of LoRAs and additional parameters trained on the DialogSum dataset by testing them on the SAMSum dataset \cite{gliwa2019samsum}. Our aim is to assess whether our method maintains strong performance and improvements when applied to data from a different domain. Specifically, DialogSum features spoken-language conversations from daily life scenarios, while SAMSum includes written-style dialogues from online interactions. For instance, DialogSum contains sentences like ``\textit{\#Person\_1\#: Good morning. I wonder whether you have got an answer from your superior.},'' whereas SAMSum includes examples such as ``\textit{Leo: BTW what are those pics?}''

Table~\ref{tab:ood} presents results for both in-domain and out-of-domain settings using the cross-lingual English-to-Spanish summarization task. The evaluation spans all three models considered earlier. The results demonstrate that the trained LoRAs and additional parameters deliver strong performance even under mild domain shifts, such as variations in conversation style.

\begin{table}[h!]
\caption{\textbf{Comparison of in-domain and out-of-domain settings.} 
Test scores reported as \% ($\uparrow$) and averaged across models for the cross-lingual English-to-Spanish summarization task. The approaches work well even in the presence of mild domain shift.}
\label{tab:ood}
\begin{center}
\resizebox{\columnwidth}{!}{
\begin{tabular}{lccccc}
\toprule
& \multicolumn{2}{c}{\textbf{In-domain}} & \multicolumn{2}{c}{\textbf{Out-of-domain}} \\
\cmidrule(lr){2-3} \cmidrule(lr){4-5}
& R-L & LLM-J & R-L & LLM-J \\
\midrule
Zero-shot & \phantom{0}9.59 & \phantom{0}0.20 & 10.58 & \phantom{0}3.54 \\
Main-task LoRA & 14.25 & \phantom{0}3.09 & 15.97 & \phantom{0}9.89 \\
Auxiliary-task LoRA & 14.42 & \phantom{0}0.16 & 19.67 & \phantom{0}1.14 \\
Linear merge & 15.60 & \phantom{0}0.18 & 18.30 & \phantom{0}1.26 \\
\hdashline
Multi-step LoRA usage & 21.71 & 75.51 & 26.81 & 76.35 \\
Joint-expert LoRA & 23.47 & 52.51 & 25.38 & 56.33 \\
\hdashline
\ours & 27.68 & 61.58 & 27.78 & 57.26 \\
\ours++ & 30.98 & 67.98 & 28.97 & 59.91 \\
\bottomrule
\end{tabular}
}
\end{center}
\end{table}

\section{Compositional Multi-tasking with Three Tasks}
\label{sec:threeway}
We demonstrate that combinations involving three tasks can also be effectively addressed using our \ours~framework. Specifically, we evaluate the joint tasks of summarization, professional tone adjustment, and translation from English to Spanish, German, or French. We created data for this task by converting ground-truth summaries into a professional tone and then translating them into multiple languages. The same models for tone adjustment and translation were applied in this setup. Given the large number of possible combinations, we focus on one tone and main task for simplicity. Additionally, we evaluate only a subset of the most relevant approaches for this scenario.

\begin{table}[h]
\caption{\textbf{Compositional multi-tasking with three-way tasks.} Combination of summarization, professional tone adjustment, and translation to various languages. Test scores reported as \% ($\uparrow$) and averaged across languages and models. Our solution also works well for three-way compositional tasks.}
\label{tab:three-tasks}
\begin{center}
\resizebox{0.76\columnwidth}{!}{
\begin{tabular}{lccc}
\toprule
 & R-L & LLM-J  \\
\midrule
Zero-shot & \phantom{0}7.35 & \phantom{0}0.03 \\
Summarization LoRA & \phantom{0}9.41 & \phantom{0}0.00 \\
Translation LoRA & 10.95 & \phantom{0}0.00 \\
Tone adj.\ LoRA & \phantom{0}8.39 & \phantom{0}0.01 \\
Linear merge & \phantom{0}9.89 & \phantom{0}0.01 \\
\hdashline
Multi-step LoRA usage & 16.50 & 43.82 \\
Joint-expert LoRA & 14.35 & 12.75 \\
\hdashline
\ours & 19.19 & 28.66 \\
\ours++ & 21.42 & 39.16 \\
\bottomrule
\end{tabular}
}
\end{center}
\end{table}

We report the results of our compositional multi-tasking analysis with three-way tasks in Table~\ref{tab:three-tasks}. The findings show: 1) inefficient baselines outperform simple baselines in this more challenging scenario, 2) our \ours~approach outperforms the inefficient baselines both in terms of accuracy and footprints, and 3) \ours++ performs better than its simpler variant. 
While the overall performance is lower than that for the simpler combination of summarization and translation, these results confirm our framework can successfully handle compositions involving more than two tasks, further proving its flexibility.

\section{Qualitative Analysis}
\label{sec:qualitative}
ROUGE scores measure the overlap between generated and reference texts but can be challenging to interpret.
To better understand these differences, we conducted a qualitative analysis.  

After inspecting model outputs and ground-truth references, we have observed there are essentially two behavioral groups. 
The first group typically does not perform one of the tasks, \eg, sometimes generates a summary but fails to translate it correctly or provides a translated text but does not summarize. This group includes approaches such as zero-shot and existing merging strategies.
The second group succeeds in performing both tasks. For instance, they correctly execute both translation and summarization.
Representatives of this category are the two inefficient baselines as well as our \ours~solutions.
The qualitative differences in ROUGE scores reflect these behaviors. Within the group itself, higher scores indicate better execution quality and consistency across more examples. The frequency with which approaches successfully complete both tasks aligns with our LLM judge analysis, providing a quantitative measure of task success.

More specifically, we perform the qualitative analysis using test examples with ROUGE scores close to the average for each method. We compare the performance of three approaches—zero-shot, main-task LoRA, and \ours++—to illustrate varying levels of success. For this analysis, we use the Qwen2.5 1.5B model with either Spanish as the target language or professional tone as the target tone.

Figures~\ref{fig:analysis-sum1}, \ref{fig:analysis-sum2} and~\ref{fig:analysis-reply} highlight common issues encountered by zero-shot and main-task LoRA methods:
\begin{enumerate}
    \item Failing to summarize the input text when summarization is required.
    \item Not adhering to the target language or tone specified in the task.
\end{enumerate}

In contrast, \ours++ demonstrates significantly better performance, successfully executing both tasks in the compositional task.

We conducted further qualitative analysis by examining outputs from different methods for the same input. This analysis focused on the cross-lingual summarization task, with results shown in Figures~\ref{fig:analysis-fixed-sample}, \ref{fig:analysis-fixed-sample2}, \ref{fig:analysis-fixed-sample3}, and~\ref{fig:analysis-fixed-sample4}.

The results reveal that multi-step and joint-expert approaches, as well as our \ours~solutions, consistently succeed in performing the compositional task. Conversely, other approaches often fail to execute one or both tasks, reinforcing the advantage of our methods in handling compositional multi-tasking scenarios.

\begin{figure*}[t]
    \centering
\begin{tcolorbox}[width=\textwidth,colback=orange!5!white, colframe=orange!75!black, title=\textbf{Cross-Lingual (English to Spanish) Summarization}]

\textit{\textbf{Zero-shot:}}

\textbf{Reference:} \#Person1\# está rogando a Steven que persuada a su esposa para que no se divorcie de él. Steven está de acuerdo.

\textbf{Prediction:} \#Persona1\#: Steven, necesito tu ayuda muy necesitado.

\textbf{Translated reference:} \#Person1\# is begging Steven to persuade his wife not to divorce him. Steven agrees.

\textbf{Translated prediction:} \#Person1\#: Steven, I need your much needed help.

\textbf{ROUGE-L score:} 7.14
\\

\textit{\textbf{Main-task LoRA:}}

\textbf{Reference:} \#Person1\# sospecha que \#Person2\# tiene varicela y quiere mantenerse alejado de \#Person2\#. \#Person2\# piensa que \#Person1\# está exagerando.

\textbf{Prediction:} \#Person1\#: ¿Qué te pasa? ¿Por qué estás picando tanto?

\textbf{Translated reference:} \#Person1\# suspects that \#Person2\# has chickenpox and wants to stay away from \#Person2\#. \#Person2\# thinks that \#Person1\# is exaggerating.

\textbf{Translated prediction:} \#Person1\#: What's wrong with you? Why are you stinging so much?

\textbf{ROUGE-L score:} 14.29
\\

\textit{\textbf{\ours++:}}

\textbf{Reference:} Frank describe el calendario de su nuevo trabajo y el seguro en la Oficina de Correos a Judy.

\textbf{Prediction:} Frank le dice a Judy que obtuvo un nuevo trabajo y le dice a Judy que el gobierno ofrece beneficios de salud de calidad.

\textbf{Translated reference:} Frank describes the schedule of his new job and insurance at Judy's Post Office.

\textbf{Translated prediction:} Frank tells Judy that he got a new job and tells Judy that the government offers quality health benefits.

\textbf{ROUGE-L score:} 28.57

\end{tcolorbox}

\caption{\textbf{Analysis of representative examples for cross-lingual summarization.} Zero-shot and main-task LoRA approaches often fail to perform summarization, instead continuing the conversation in Spanish. In contrast, our approach successfully executes both summarization and translation or tone adjustment tasks.}
\label{fig:analysis-sum1}
\end{figure*}

\begin{figure*}[t]
    \centering

\begin{tcolorbox}[width=\textwidth,colback=orange!5!white, colframe=orange!75!black, title=\textbf{Cross-Tone (Professional) Summarization}]

\textit{\textbf{Zero-shot:}}

\textbf{Reference:} \#Person1\# is not satisfied with the steak and is considering alternatives.

\textbf{Prediction:} \#Person1\#: Excuse me.

\textbf{ROUGE-L score:} 14.29
\\

\textit{\textbf{Main-task LoRA:}}

\textbf{Reference:} In order to prevent employees from engaging in unnecessary activities, Mr. \#1\# decides to discontinue the use of Instant Messaging programs and requests Ms. Dawson to issue a memo to all employees by the morning.

\textbf{Prediction:} \#Person1\#: Ms. Dawson, I need you to take a dictation for me.

\textbf{ROUGE-L score:} 17.02
\\

\textit{\textbf{\ours++:}}

\textbf{Reference:} As a result of \#Person1's desire to visit the science museum, they became lost. To resolve the issue, \#Person2\# assisted them in purchasing a ticket and providing them with directions.

\textbf{Prediction:} \#Person2\# assists \#Person1\# in navigating the science museum and provides guidance on the ticket machines.

\textbf{ROUGE-L score:} 26.09

\end{tcolorbox}

\caption{\textbf{Analysis of representative examples for cross-tone summarization.} Zero-shot and main-task LoRA approaches often fail to perform summarization, instead continuing the conversation in Spanish. In contrast, our approach successfully executes both summarization and translation or tone adjustment tasks.}
\label{fig:analysis-sum2}
\end{figure*}

\begin{figure*}[t]
    \centering
\begin{tcolorbox}[width=\textwidth,colback=orange!5!white, colframe=orange!75!black, title=\textbf{Cross-Lingual (English to Spanish) Reply Suggestion}]

\textit{\textbf{Zero-shot:}}

\textbf{Reference:} Enseño biología en una universidad.

\textbf{Prediction:} What do you do for a living?

\textbf{Translated reference:} I teach biology at a university.

\textbf{Translated prediction:} What do you do for a living?

\textbf{Weighted ROUGE score:} 2.38
\\

\textit{\textbf{Main-task LoRA:}}

\textbf{Reference:} Sólo he hecho modelaje de moda.

\textbf{Prediction:} Oh, that's cool! I've always wanted to be a model.

\textbf{Translated reference:} I've only done fashion modeling.

\textbf{Translated prediction:} Oh, that's cool! I've always wanted to be a model.

\textbf{Weighted ROUGE score:} 1.75
\\

\textit{\textbf{\ours++:}}

\textbf{Reference:} Me gusta ir al cine, jugar videojuegos y pasar el rato con mis amigos.

\textbf{Prediction:} Me gusta pasar tiempo con mi familia y amigos.

\textbf{Translated reference:} I like to go to the movies, play video games and hang out with my friends.

\textbf{Translated prediction:} I like spending time with my family and friends.

\textbf{Weighted ROUGE score:} 11.87

\end{tcolorbox}

\begin{tcolorbox}[width=\textwidth,colback=orange!5!white, colframe=orange!75!black, title=\textbf{Cross-Tone (Professional) Reply Suggestion}]

\textit{\textbf{Zero-shot:}}

\textbf{Reference:} Yes, it is indeed!

\textbf{Prediction:} Absolutely, it sounds like a fantastic opportunity!

\textbf{Weighted ROUGE score:} 3.03
\\

\textit{\textbf{Main-task LoRA:}}

\textbf{Reference:} It's an exciting opportunity to unleash your creativity and unleash the power of apps and websites.

\textbf{Prediction:} It's a lot of fun! You get to create your own programs and apps.

\textbf{Weighted ROUGE score:} 9.51
\\

\textit{\textbf{\ours++:}}

\textbf{Reference:} I enjoy attending the movies and engaging in leisure activities such as reading.

\textbf{Prediction:} I enjoy reading, watching movies, and spending time with my family.

\textbf{Weighted ROUGE score:} 13.01

\end{tcolorbox}

\caption{\textbf{Analysis of representative examples for cross-lingual and cross-tone reply suggestion.} Zero-shot and main-task LoRA approaches often fail to use the correct language or apply the desired tone. In contrast, our approach successfully executes both reply suggestion and translation or tone adjustment tasks.}
\label{fig:analysis-reply}
\end{figure*}

\begin{figure*}[t]
    \centering
\begin{tcolorbox}[width=\textwidth,colback=orange!5!white, colframe=orange!75!black, title=\textbf{Cross-Lingual (English to Spanish) Summarization}]
\textbf{Reference:} Eric y Rob van a ver un stand-up en YouTube.

\textbf{Translated reference:} Eric and Rob are going to watch a stand-up on YouTube.\\

\textbf{Zero-shot prediction:} Eric: Máquina!

\textbf{Zero-shot translated prediction:} Eric: Machine!\\

\textbf{Main-task LoRA prediction:} Eric: Máquina!

\textbf{Main-task LoRA translated prediction:} Eric: Machine!\\

\textbf{Auxiliary-task LoRA prediction:} Eric: Máquina! Rob: ¡Esto es genial! Eric: Lo sé. Y muestra cómo los estadounidenses ven a Rusia. Rob: Y es muy divertido. Eric: Lo sé. Y me encanta la parte del tren. Rob: ¡Hahahaha! No se habla a la máquina así. Eric: ¿Y es su única actuación? Rob: No sé. Lo veré. Eric: Sí. Rob: ¡Sí!

\textbf{Auxiliary-task LoRA translated prediction:} Eric: Machine! Rob: This is great! Eric: I know. And it shows how Americans see Russia. Rob: And it’s really fun. Eric: I know. And I love the part of the train. Rob: Hahaha! You don’t talk to the machine like that. Eric: And it’s your only performance? Rob: I don’t know. I’ll see it. Eric: Yes. Rob: Yes!\\

\textbf{Linear merge prediction:} Eric: Máquina!

\textbf{Linear merge translated prediction:} Eric: Machine!\\

\textbf{Multi-step LoRA usage prediction:} Eric y Rob hablan de un video divertido que muestra una máquina hablando como estadounidenses. Lo encuentran divertido y planean ver más de los videos de stand-up de Eric.

\textbf{Multi-step LoRA usage translated prediction:} Eric and Rob talk about a funny video that shows a machine talking like Americans. They find it fun and plan to see more of Eric's stand-up videos.\\

\textbf{Joint-expert LoRA prediction:} Eric y Rob están viendo un stand-up de Eric, un hombre americano que hace stand-up. Eric está muy divertido y se ríe mucho. Rob también está viendo y se ríe.

\textbf{Joint-expert LoRA translated prediction:} Eric and Rob are watching a stand-up by Eric, an American man who makes stand-up. Eric is very funny and laughs a lot. Rob is also watching and laughing.\\

\textbf{\ours~prediction:} Rob y Eric están interesados en ver los vídeos de stand-up de Eric.

\textbf{\ours~translated prediction:} Rob and Eric are interested in watching Eric's stand-up videos.\\

\textbf{\ours++ prediction:} Eric y Rob están viendo los stand-up de Eric.

\textbf{\ours++ translated prediction:} Eric and Rob are watching Eric's stand-up.\\

\end{tcolorbox}

\caption{\textbf{Analysis of representative examples for cross-lingual summarization using a fixed input.}}
\label{fig:analysis-fixed-sample}
\end{figure*}

\begin{figure*}[t]
    \centering
\begin{tcolorbox}[width=\textwidth,colback=orange!5!white, colframe=orange!75!black, title=\textbf{Cross-Tone (Professional) Summarization}]
\textbf{Reference:} \#Person2\# was late due to traffic jam. \#Person1\# encouraged \#Person2\# to utilize public transportation to maintain a healthy lifestyle and safeguard the environment.\\

\textbf{Zero-shot prediction:} \#Person1\#: You're finally here! What took so long?\\

\textbf{Main-task LoRA prediction:} \#Person2\# is going to quit driving to work.\\

\textbf{Auxiliary-task LoRA prediction:} Summary: Person 2 is stuck in traffic again, and Person 1 suggests finding a different route or using public transport. Person 2 acknowledges the benefits of public transport but is concerned about missing the freedom of driving. Person 1 suggests biking to work when the weather is nicer, which Person 2 agrees with. Person 2 decides to quit driving to work and switch to public transport or biking.\\

\textbf{Linear merge prediction:} \#Person1\#: You're finally here! What took so long?\\

\textbf{Multi-step LoRA usage prediction:} The individual designated as \#Person2\# will be ceasing their practice of commuting via automobile for employment purposes.\\

\textbf{Joint-expert LoRA prediction:} \#Person2\# is considering switching to public transport to work.\\

\textbf{\ours~prediction:} \#Person2\# arrived late due to traffic congestion, and \#Person1\# suggested that \#Person2\# consider taking public transport to work.\\

\textbf{\ours++ prediction:} \#Person2\# is stuck in traffic and is considering taking public transport to work. \#Person1\# suggests that \#Person2\# consider biking to work.\\
\end{tcolorbox}

\caption{\textbf{Analysis of representative examples for cross-tone summarization using a fixed input.}}
\label{fig:analysis-fixed-sample2}
\end{figure*}

\begin{figure*}[t]
    \centering
\begin{tcolorbox}[width=\textwidth,colback=orange!5!white, colframe=orange!75!black, title=\textbf{Cross-Lingual (English to Spanish) Reply Suggestion}]
\textbf{Reference:} ¡Lo es!\\
\textbf{Translated reference:} It is!\\

\textbf{Zero-shot prediction:} ¡Eso suena como un trabajo excelente!

\textbf{Zero-shot translated prediction:} That sounds like an excellent job!\\

\textbf{Main-task LoRA prediction:} Thanks! I'm really enjoying it.

\textbf{Main-task LoRA translated prediction:} Thank you! I'm really enjoying it.\\

\textbf{Auxiliary-task LoRA prediction:} Eso suena como un trabajo excelente!

\textbf{Auxiliary-task LoRA translated prediction:} That sounds like an excellent job!\\

\textbf{Linear merge prediction:} ¡Eso suena genial!

\textbf{Linear merge translated prediction:} That sounds great!\\

\textbf{Multi-step LoRA usage prediction:} Sí, me encanta.

\textbf{Multi-step LoRA usage translated prediction:} Yeah, I love it.\\

\textbf{Joint-expert LoRA prediction:} Me encanta trabajar con niños.

\textbf{Joint-expert LoRA translated prediction:} I love working with kids.\\

\textbf{\ours~prediction:} Sí, es muy divertido.

\textbf{\ours~translated prediction:} Yeah, it's a lot of fun.\\

\textbf{\ours++ prediction:} Gracias, lo es.

\textbf{\ours++ translated prediction:} Thank you, it is.
\end{tcolorbox}
\caption{\textbf{Analysis of representative examples for cross-lingual reply suggestion using a fixed input.}}
\label{fig:analysis-fixed-sample3}
\end{figure*}

\begin{figure*}[t]
    \centering
\begin{tcolorbox}[width=\textwidth,colback=orange!5!white, colframe=orange!75!black, title=\textbf{Cross-Tone (Professional) Reply Suggestion}]
\textbf{Reference:} I will certainly consider your recommendation. Thank you for the valuable input.\\

\textbf{Zero-shot prediction:} Absolutely, it is a stunning location.\\

\textbf{Main-task LoRA prediction:} I will! Thanks for the recommendation.\\

\textbf{Auxiliary-task LoRA prediction:} Absolutely, it's a wonderful location.\\

\textbf{Linear merge prediction:} I agree. It's a great place to visit.\\

\textbf{Multi-step LoRA usage prediction:} I will! Thank you for your recommendation.\\

\textbf{Joint-expert LoRA prediction:} I will certainly consider it. Thank you for the recommendation.\\

\textbf{\ours~prediction:} I will definitely consider it.\\

\textbf{\ours++ prediction:} I will consider it.
\end{tcolorbox}
\caption{\textbf{Analysis of representative examples for cross-tone reply suggestion using a fixed input.}}
\label{fig:analysis-fixed-sample4}
\end{figure*}

\section{Learning Curves Analysis of Our Approach}
\label{sec:learning-curves}
\begin{figure*}[ht]
\vskip 0.2in
\begin{center}
\centerline{\includegraphics[width=\linewidth]{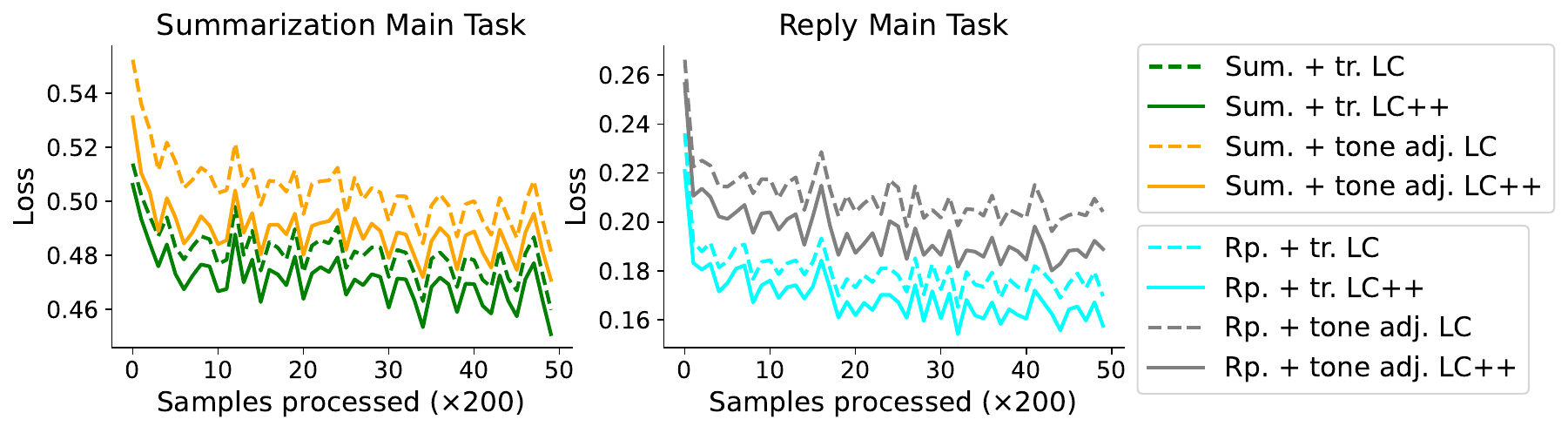}}
\caption{\textbf{Learning curves analysis of our approach.} 
The training loss for \ours~parameters decreases consistently as training progresses.}
\label{fig:learning_curve}
\end{center}
\vskip -0.2in
\end{figure*}

We analyze the training process of \ours~in both variations to better understand its behavior. 
For this analysis, we use the Qwen2.5 1.5B model with Spanish as the target language and a professional target tone. 
The results in Figure~\ref{fig:learning_curve} shows the loss consistently decreases as training progresses and more samples are processed.

\end{document}